\documentclass{article}

\usepackage{arxiv}

\usepackage[utf8]{inputenc}
\usepackage[T1]{fontenc}
\usepackage{hyperref}
\usepackage{url}
\usepackage{booktabs}
\usepackage{amsfonts}
\usepackage{amsmath}
\usepackage{amssymb}
\usepackage{makecell}
\usepackage{microtype}
\usepackage{graphicx}
\usepackage{natbib}
\usepackage{doi}
\usepackage{array}
\usepackage{multirow}
\usepackage{placeins}
\usepackage{tabularx}

\newcolumntype{Y}{>{\raggedright\arraybackslash}X}

\title{Efficient INT8 Inference of Small NLP Models on Server CPUs with PyTorch Native Stack}

\author{
  Weiwen Xia, Yuxin Cui, E Cao \\
  Intel Corporation \\
  \texttt{\{weiwen.xia,lily.cui,e.cao\}@intel.com}
}
\date{}

\renewcommand{\shorttitle}{INT8 Inference of Small NLP Models on Xeon CPUs}

\hypersetup{
  pdftitle={Efficient INT8 Inference of Small NLP Models on Server CPUs with PyTorch Native Stack},
  pdfauthor={Weiwen Xia, Yuxin Cui, E Cao},
  pdfkeywords={NLP, BERT, INT8 Quantization, SmoothQuant, PyTorch, CPU inference, Roofline Model, Intel Xeon}
}

\begin{document}
\maketitle

\begin{abstract}
Small NLP models, especially BERT-family encoders, remain important in industrial workloads such as classification, ranking, and retrieval even in the era of large language models. On server CPUs, INT8 quantization offers an attractive latency-throughput-cost trade-off, but users increasingly expect such acceleration to be available directly in the native PyTorch stack. We integrate SmoothQuant into TorchAO and optimize the resulting inference path for Intel Xeon CPUs through graph-level fusion in TorchInductor and efficient INT8 GEMM kernel selection across oneDNN-, AVX512\_VNNI-, and AMX-based implementations. Across BERT, DistilBERT, and XLM-RoBERTa benchmarks, the approach delivers up to $5.8\times$ end-to-end throughput speedup with negligible---and in some cases no measurable---accuracy loss relative to the FP32 baseline. We also validated our work by detailed performance analysis with roofline models. The implementation has been upstreamed to PyTorch and TorchAO, enabling out-of-the-box deployment with native PyTorch tooling.
\end{abstract}


\section{Introduction}

Although large language models (LLMs) have attracted much of the recent attention, small NLP models based on the BERT family remain widely deployed for production tasks such as classification, ranking, and retrieval \citep{devlin2019bert,sanh2019distilbert,conneau2020unsupervised}. CPU inference continues to play a central role in such deployments because of its favorable cost profile and the ubiquity of server CPU infrastructure.

Within the PyTorch ecosystem, there is strong demand for efficient inference using the native software stack. Straightforward FP32 execution on CPU is often suboptimal because transformer inference is dominated by matrix multiplications (GEMM) with substantial compute and memory cost. INT8 quantization is a widely adopted technique for reducing both footprint and execution time while maintaining acceptable accuracy \citep{jacob2018quantization,nagel2021whitepaper}. At the same time, modern Intel Xeon CPUs expose instruction-set support such as VNNI and AMX that can substantially accelerate INT8 GEMM workloads \citep{intel_isa_extensions}.

In this work, we integrate SmoothQuant \citep{xiao2023smoothquant}, a post-training quantization method designed to mitigate activation outliers in transformers, into the native PyTorch stack and optimize its performance on modern Xeon CPUs. Our contributions are:

\begin{itemize}
  \item End-to-end SmoothQuant workflow is integrated into TorchAO's frontend \citep{or2025torchao} and improved for subsequent graph optimizations.
  \item Graph fusion passes in TorchInductor \citep{ansel2024pytorch2} is added for INT8 GEMM operators eliminating runtime overheads of weight layout conversion and handling of post-operations.
  \item Kernel selection is enabled for INT8 GEMM for peak performance to choose between oneDNN-backed kernels \citep{onednn} and template-based GEMM kernels in TorchInductor's CPP backend.
  \item AVX512\_VNNI microkernel is added in TorchInductor's CPP backend to cover 3rd Gen Xeon CPUs in addition to AMX-based kernels for newer platforms.
  \item Comprehensive performance analysis with roofline models and benchmarks are conducted to show the effectiveness of the proposed workflow and optimizations.
  \item Benchmarks are conducted on BERT, DistilBERT, and XLM-RoBERTa over SQuAD \citep{rajpurkar2016squad} and MultiNLI \citep{williams2018mnli} showing negligible accuracy loss and substantial throughput gains compared to FP32 baselines.
\end{itemize}

This work strengthens the PyTorch-native inference path for small NLP models on server CPUs, enabling users to deploy SmoothQuant-optimized INT8 inference with minimal effort and without leaving the PyTorch ecosystem or using third-party tools.

\section{Related Work}

Quantization has long been studied as a way to improve inference efficiency. Integer-only quantization was popularized by work such as \citet{jacob2018quantization}, and later surveys summarize the broader landscape of post-training quantization (PTQ) and quantization-aware training (QAT) \citep{nagel2021whitepaper}.

For transformer models, several methods explicitly target quantization difficulty caused by activation outliers or model sensitivity. LLM.int8() \citep{dettmers2022llm8} introduces mixed-precision handling for outlier channels in LLMs. GPTQ \citep{frantar2023gptq} focuses on accurate weight-only post-training quantization, and AWQ \citep{lin2023awq} uses activation-aware weight quantization. Compared with these methods, SmoothQuant \citep{xiao2023smoothquant} shifts activation difficulty into the weights so that standard INT8 arithmetic can be applied efficiently on common hardware.

Several software stacks support quantized inference on CPUs. Intel Neural Compressor \citep{intel_neural_compressor} provides broad quantization support but typically relies on PyTorch itself or other third-party tools for deployment. Hugging Face Optimum Intel \citep{optimum_intel} integrates Intel-oriented optimization into the Hugging Face ecosystem, while ONNX Runtime \citep{onnxruntime} provides INT8 inference through exported ONNX graphs. SGLang \citep{zheng2024sglang} and vLLM \citep{kwon2023vllm} are important LLM-serving systems, but they focus on decoder-style LLM serving rather than BERT-family encoder inference and do not target SmoothQuant in native PyTorch. Intel Extension for PyTorch (IPEX) \citep{intel_ipex} previously offered an efficient SmoothQuant workflow on top of oneDNN, but that path relied on legacy \texttt{torch.jit}-based optimization; its maintenance has since shifted toward native PyTorch 2.x compilation and upstream integration.

Our work differs from these approaches by delivering a fully native PyTorch solution that combines SmoothQuant, TorchAO, TorchInductor graph fusion, and ISA-aware CPU kernel selection in one workflow. This integration allows for seamless deployment of optimized INT8 inference within the PyTorch ecosystem.

\section{Background and Challenges}

\subsection{SmoothQuant and Quantization Modes}

Activation outliers are common in transformer models and can lead to large quantization error under straightforward INT8 quantization. SmoothQuant \citep{xiao2023smoothquant} addresses this issue by redistributing scale factors between activations and weights: the activation tensor is divided by a per-channel smoothing factor, while the corresponding weight tensor is multiplied by the same factor. This transformation reduces activation outliers while preserving the linear layer mathematically.

The smoothing factors are derived offline using a small calibration dataset. After smoothing, the model becomes substantially more amenable to standard INT8 quantization \citep{jacob2018quantization,nagel2021whitepaper}.

The resulting quantization flow can be either static or dynamic, depending on when activation quantization parameters are determined. In static quantization, scales and zero points are obtained during calibration and then fixed in the graph. In dynamic quantization, they are computed at runtime and may vary by input. Static quantization generally offers lower runtime overhead, whereas dynamic quantization can sometimes preserve accuracy more effectively at the cost of additional execution work.

\subsection{PyTorch and TorchAO}

PyTorch is widely used for both research and production \citep{paszke2019pytorch}. PyTorch 2 introduces TorchInductor (its API \texttt{torch.compile} might be better known), which enables graph capture, optimization, and backend lowering for improved execution efficiency \citep{ansel2024pytorch2}. On CPU, TorchInductor's CPP backend is a key path for native server-class inference performance. Ahead-of-Time Inductor (AOTI) is a specialized version of TorchInductor that compiles the model into a shared binary, eliminating Python overhead and enabling easier integration into production systems.

TorchAO is the PyTorch-native quantization library that provides user-friendly APIs for PTQ and QAT \citep{or2025torchao}. It supports a range of low-precision data types and quantization methods through a modular design.

\subsection{Challenges}

The integration of SmoothQuant is straightforward in terms of correctness, but achieving high performance on CPU requires additional engineering effort. TorchAO quantizes models in eager mode without backend-aware lowering or graph-level optimization. That makes the resulting model correct, but not necessarily fast enough for production CPU inference. To achieve peak performance with native PyTorch, the quantized model must be compiled with TorchInductor, which introduces several engineering challenges:

\begin{itemize}
  \item The SmoothQuant workflow shares common code with existing TorchAO quantization flows, which contain inefficient operations for high-performance inference.
  \item The scaled INT8 GEMM operator is convenient for modeling and prototyping but has limited performance.
  \item Leveraging TorchInductor's kernel-selection mechanism is a key strategy so that different shapes can choose the best backend implementation.
  \item Supporting both legacy AVX512\_VNNI-only platforms, such as Ice Lake, and newer AMX-capable Xeon generations.
\end{itemize}

\section{Design and Implementation}

\subsection{Overall Design}

Our workflow is built on top of TorchAO's quantization path and TorchInductor's graph optimization and CPP backend. To apply SmoothQuant to a pretrained model, the user invokes TorchAO's high-level API, which:

\begin{itemize}
  \item prepares the model for calibration by inserting observers for activation statistics;
  \item searches for smoothing factors by running calibration data through the prepared model;
  \item replaces \texttt{torch.nn.Linear} weights with a custom tensor subclass that stores quantized weights, scales, and smoothing factors; and
  \item replaces high-precision linear operations with INT8 $\times$ INT8 $\rightarrow$ INT32 GEMM via \texttt{torch.\_int\_mm}, followed by output scaling.
\end{itemize}

After quantization, linear layers execute in INT8 while the remaining operators stay in higher precision. The model is then compiled through TorchInductor, which applies graph optimizations and lowers the result into the CPP backend. We added graph passes that fuse scaled INT8 GEMM subgraphs centered on \texttt{torch.\_int\_mm} into the oneDNN-backed operator \texttt{torch.ops.onednn.qlinear\_pointwise}. We also extended TorchInductor's template-based GEMM path with AMX and AVX512\_VNNI microkernels so that lowering can select between the oneDNN-based kernel and generated template kernels.

\begin{figure}[t]
  \centering
  \includegraphics[width=\textwidth]{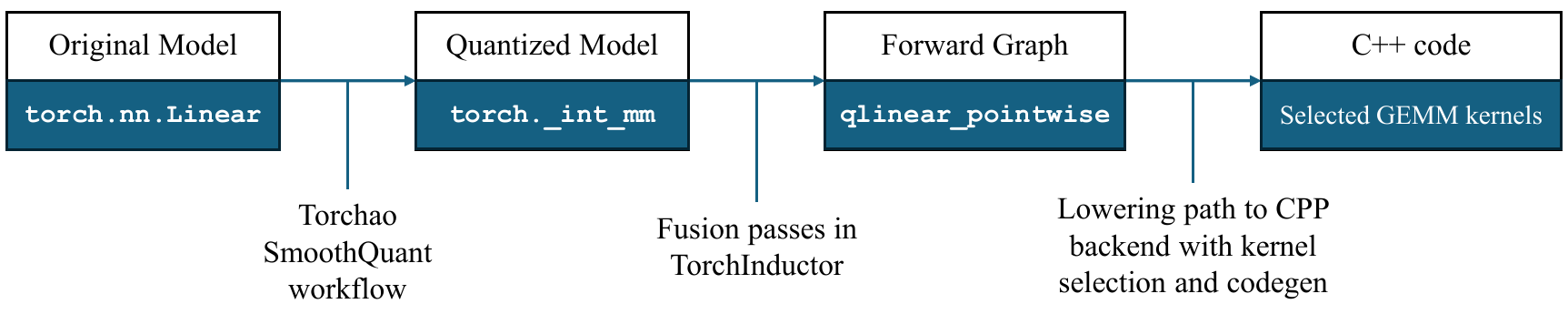}
  \caption{SmoothQuant workflow built on TorchAO and TorchInductor. TorchAO prepares and quantizes the model, TorchInductor fuses INT8 subgraphs, and the CPP backend selects between oneDNN and template-based GEMM kernels. C++ code is generated and compiled for deployment.}
  \label{fig:workflow}
\end{figure}

\subsection{Improving Implementation of Common Quantization Flows}

The integration of SmoothQuant shares much common code with existing TorchAO quantization flows, and we made several improvements to improve runtime performance:

\begin{itemize}
  \item When applying the smoothing factor, we precompute the inverse of the factor to avoid repeated division at runtime.
  \item We avoid tensor copies by ignoring a few unnecessary tensor expand operations.
  \item We ignore dummy zero points to save computation.
\end{itemize}

\subsection{Overcoming \texttt{torch.\_int\_mm} Limitations}
\label{sec:intmm-limitations}

We implemented the CPU version of \texttt{torch.\_int\_mm} using oneDNN to support INT8 GEMM for cases like the SmoothQuant flow in TorchAO. However, the primitive by itself introduces three important limitations:

\begin{itemize}
  \item \textbf{Runtime weight layout conversion.} Efficient INT8 GEMM on Xeon often requires blocked weight layouts tailored for VNNI and AMX instructions. Converting weights from plain layout to blocked layout at runtime is expensive.
  \item \textbf{Missing scale and post-operation handling.} The operator computes INT8 $\times$ INT8 $\rightarrow$ INT32 GEMM without an epilogue, so scales and other post-operations need to be handled separately.
  \item \textbf{oneDNN setup overhead.} oneDNN requires auxiliary data structures whose per-call construction can become nontrivial overhead.
\end{itemize}

We address these issues by extending the graph fusion capability in TorchInductor. The following pattern and its variants are matched:

\begin{center}
\texttt{input\_reshape -> int\_mm -> dtype\_convert -> mul -> mul -> output\_reshape -> add\_bias}
\end{center}

where \texttt{int\_mm} is the INT8 GEMM operator, the first \texttt{mul} applies the activation scale, the second \texttt{mul} applies the weight scale, and \texttt{add\_bias} adds a bias term. The matched subgraph is then replaced by a single \texttt{torch.ops.onednn.qlinear\_pointwise} operator. This fused operator accepts blocked-layout weights, applies activation and weight scales as post-operations inside the kernel, and caches oneDNN metadata after warm-up. This kernel can be further fused with other operators such as \texttt{GeLU} \citep{hendrycks2016gelu} by existing fusion capability of TorchInductor. A corresponding weight layout conversion operator is also added to the graph to convert weights from plain layout to blocked layout, which is evaluated and folded during a constant folding pass in TorchInductor.

\subsection{Template-Based GEMM Kernels for Maximum Performance}

Although \texttt{qlinear\_pointwise} performs well in many cases, it is not uniformly optimal across all matrix shapes. We therefore extended TorchInductor's template-based GEMM kernels as an alternative lowering target. Specifically, we implemented INT8 GEMM microkernels using AMX and AVX512\_VNNI intrinsics and integrated them into the CPP backend's generated kernel path.

During lowering, TorchInductor's kernel-selection mechanism generates candidate kernels, compiles them, micro-benchmarks them, and selects the fastest implementation for future execution. This one-time selection process enables the system to balance the generality of oneDNN against the shape specialization of template-generated kernels.

\subsection{Supporting Ice Lake: $s8s8 \rightarrow u8s8$ Transformation}

TorchAO quantizes both activations and weights to signed INT8 ($s8$). However, 3rd Gen Xeon (Ice Lake) does not provide native $s8 \times s8$ GEMM instructions; AVX512\_VNNI requires activation to be unsigned INT8 $u8$, i.e. $u8 \times s8$. For GEMM of $Y = X \cdot W^\top$, we therefore shift the activation tensor $X$ by a constant:

\begin{equation}
  X_{u8} = \operatorname{uint8}(\operatorname{int32}(X_{s8}) + 128).
  \label{eq:u8shift}
\end{equation}

To preserve mathematical equivalence, the shift is treated as a zero point and compensated in the GEMM result. For a weight matrix $W \in \mathbb{Z}^{N \times K}$, the compensation term is

\begin{equation}
  c = \left(\sum_{k=1}^{K} W_{:,k}\right) \cdot 128.
  \label{eq:compensation}
\end{equation}

And the GEMM result is computed as
\begin{equation}
  Y = X_{u8} \cdot W^\top - c.
  \label{eq:gemm-compensation}
\end{equation}

Because both the shift and the weights are constant, the weight conversion and compensation vector are precomputed before inference. Although this introduces some overhead for loading and subtraction, it still yields substantial speedup by enabling AVX512\_VNNI execution on Ice Lake. The fusion pattern in Section~\ref{sec:intmm-limitations} is extended to include the shift and compensation accordingly.

\section{Performance Analysis and Optimization Targets}
\label{sec:perf-analysis}

\subsection{Benchmark Setup for Baseline Collection}

BERT-family models contain both the GEMM-heavy layers that benefit most from quantization and the higher-precision operators that still limit end-to-end speedup and they are representative of the small NLP models that remain widely deployed in production. To ground the performance discussion in a concrete baseline, we begin with a focused study that identifies the dominant execution hotspots and quantifies the headroom for INT8 acceleration.

We conduct benchmarks on two representative Xeon platforms: 3rd Gen (Ice Lake) and 6th Gen (Granite Rapids). Table~\ref{tab:hardware} summarizes the hardware platforms. The 4th and 5th Gen Xeon platforms share the same INT8 GEMM instruction support as the 6th Gen architecture, so we omit them for scope; ISA-level conclusions transfer, while absolute performance may differ across generations.

For the performance benchmark, we measure end-to-end throughput in requests per second (RPS) and linear-block latency (GEMM + fused post-operations) using synthetic inputs with batch size 1. Table~\ref{tab:baseline-config} summarizes the baseline collection setup. The multi-instance launch strategy is used to maximize throughput by running multiple independent model instances in parallel, each pinned to a subset of physical cores: 16 instances on 64 physical cores for Ice Lake, and 64 instances on 256 physical cores for Granite Rapids. Instances are partitioned per socket; within each socket, instances share model weights and synthetic inputs to maximize utilization of last level cache (LLC) and avoid cross-socket memory access. We collect FP32 baselines on both evaluated platforms and BF16 on Granite Rapids, where native BF16 compute is available. For reproducibility, (1) each instance is pinned with \verb|taskset| to physical cores only, with \verb|OMP_NUM_THREADS=4|, \verb|KMP_AFFINITY=granularity=fine,compact,1,<first_core>|, and \verb|torch.set_num_threads(4)|; (2) Intel OpenMP runtime library and the TCMalloc memory allocator are used; (3) the first 100 iterations are warm-up and the next 200 iterations are measured; (4) the total throughput and latency are reported.

\begin{table}[t]
  \centering
  \caption{Hardware platforms used for baseline collection.}
  \label{tab:hardware}
  \begin{tabularx}{\textwidth}{l l l l Y}
    \toprule
    Xeon Generation & Codename & Model & Physical Cores & Instruction Support \\
    \midrule
    3rd Gen & Ice Lake & 8358 & 64 & AVX512F, AVX512\_VNNI \\
    6th Gen & Granite Rapids & 6980P & 256 & AVX512F, AVX512\_VNNI, AMX \\
    \bottomrule
  \end{tabularx}
\end{table}

\begin{table}[t]
  \centering
  \caption{Baseline collection setup.}
  \label{tab:baseline-config}
  \begin{tabular}{ll}
    \toprule
    Item & Value \\
    \midrule
    Model & BERT-large \\
    Input shape & Batch size 1, sequence length 256 \\
    Precision modes & FP32 on both platforms; BF16 on Granite Rapids \\
    Launch strategy & Multi-instance with weight-sharing on each socket \\
    Cores per instance & 4 \\
    Number of instances & Number of physical cores / 4 \\
    Deployment path & AOTI enabled \\
    Input data & Synthetic \\
    \bottomrule
  \end{tabular}
\end{table}

\subsection{Baseline Hotspot Breakdown}

We focus on BERT-large \citep{devlin2019bert} in this section and subsequent sections and similar analysis goes for other BERT-family models.

Table~\ref{tab:fp32-baseline-hotspots} summarizes profiler-derived operator breakdowns for the FP32 baselines on both platforms and the BF16 baseline on Granite Rapids. The baseline already reveals the core optimization target of this work. In the FP32 baselines, about $87\%$ of self CPU time on Granite Rapids and about $89\%$ on Ice Lake are spent in linear-block kernels, while attention accounts for about $12\%$ and $10\%$, respectively. For BF16 on Granite Rapids, linear-block share decreases while attention and normalization shares increase. These baselines justify treating linear-block/GEMM as the main hotspot while also making clear that non-GEMM operators limit end-to-end speedup after GEMM is accelerated. Due to profiler limitations, we cannot separate pure GEMM from linear blocks. We treat linear-block as an approximation of GEMM for the following analysis, which is reasonable because GEMM dominates linear-blocks.

\begin{table}[t]
  \centering
  \caption{Baseline hotspot breakdown for the BERT-large analysis case under the same multi-instance setup used for throughput benchmarking. Operator shares are aggregated from representative profiler logs. The linear-block consists of GEMM and fused post-operations such as bias addition, activation, and residual addition.}
  \label{tab:fp32-baseline-hotspots}
  \footnotesize
  \begin{tabular}{lrrrr}
    \toprule
    Baseline & Linear-Block & Attention & LayerNorm & Other \\
    \midrule
    FP32 on Ice Lake & 88.5\% & 10.4\% & 0.6\% & 0.5\% \\
    FP32 on Granite Rapids & 86.9\% & 11.8\% & 0.8\% & 0.5\% \\
    BF16 on Granite Rapids & 72.70\% & 19.75\% & 5.99\% & 1.56\% \\
    \bottomrule
  \end{tabular}
\end{table}

\subsection{INT8 GEMM Ceiling Estimation}

\begin{table}[t]
  \centering
  \caption{Estimated peak compute and measured memory bandwidth per socket, balance point for different data types on the evaluated platforms.}
  \label{tab:machine-balance}
  \footnotesize
  \begin{tabular}{cccccccc}
    \toprule
    Platform & \makecell{FP32 \\ Throughput \\ (TFLOPS)} & \makecell{BF16 \\ Throughput \\ (TFLOPS)} & \makecell{INT8 \\ Throughput \\ (TOPS)} & \makecell{Memory \\ Bandwidth \\ (GB/s)} & \makecell{FP32 \\ Balance Point \\ (FLOPs/byte)} & \makecell{BF16 \\ Balance Point \\ (FLOPs/byte)} & \makecell{INT8 \\ Balance Point \\ (OPs/byte)} \\
    \midrule
    Ice Lake & 6.76 & N/A & 27.03 & 175.00 & 38.63 & N/A & 154.46 \\
    Granite Rapids & 19.66 & 262.14 & 524.29 & 700.00 & 28.09 & 374.49 & 748.99 \\
    \bottomrule
  \end{tabular}
\end{table}

\begin{table}[!t]
  \centering
  \caption{BERT-large GEMM shapes and their arithmetic intensity.}
  \label{tab:bert-gemm-shapes}
  \footnotesize
  \begin{tabular}{ccccc}
    \toprule
     & & \multicolumn{3}{c}{Arithmetic Intensity (ops/byte)} \\
    \cmidrule(lr){3-5}
    \thead{GEMM Shape\\M $\times$ K $\times$ N} & \thead{Occurrences per\\forward pass} & FP32 & BF16 & INT8 \\
    \midrule
    $256 \times 1024 \times 1024$ & 96 & 85.33 & 146.29 & 227.56 \\
    $256 \times 1024 \times 4096$ & 24 & 97.52 & 163.84 & 248.24 \\
    $256 \times 4096 \times 1024$ & 24 & 97.52 & 186.18 & 341.33 \\
    \bottomrule
  \end{tabular}
\end{table}

According to the roofline model \citep{williams2009roofline}, the peak memory bandwidth and compute throughput of the evaluated platforms can be used to estimate the balance point (BP) for different data types by
\begin{equation}
  BP = \frac{P_{\text{compute}}}{B}
\end{equation}
where $P_{\text{compute}}$ is the peak compute throughput and $B$ is the peak memory bandwidth. The balance point indicates the arithmetic intensity at which a kernel transitions from being memory-bound to compute-bound. The results are listed in Table~\ref{tab:machine-balance}, where the theoretical throughputs are estimated by hardware specs and the memory bandwidth is from measurement.

The shapes of BERT-large GEMM and their arithmetic intensity (AI) are shown in Table~\ref{tab:bert-gemm-shapes}, which are evaluated by
\begin{equation}
  AI = \frac{2 \cdot M \cdot N \cdot K}{E_{in} \cdot (M \cdot K + K \cdot N) + E_{out} \cdot (M \cdot N)}
\end{equation}
where $E_{in}$ and $E_{out}$ are the data sizes of input and output elements respectively and $M$, $K$, and $N$ are the GEMM dimensions. We have disregarded the \verb`lm_head` layer for simplicity because it accounts for a negligible proportion of the time. Apparently, (1) FP32 GEMM is always compute-bound on both platforms (AI $>$ BP); (2) BF16 GEMM is memory-bound on Granite Rapids (AI $<$ BP); and (3) INT8 GEMM is compute-bound on Ice Lake (AI $>$ BP) but memory-bound on Granite Rapids (AI $<$ BP).

According to the roofline model, theoretical peak throughput of GEMM is computed by
\begin{equation}
  P_{\text{GEMM}} = \min(P_{\text{compute}}, \; B \cdot AI)
\end{equation}
where $P_{\text{compute}}$ is the peak compute throughput, $B$ the peak memory bandwidth, and $AI$ the arithmetic intensity.

Therefore, the estimated GEMM speedup on Ice Lake is $4.0\times$. And on Granite Rapids, GEMM is memory-bound for both BF16 and INT8. However, it is noted that the size of quantized INT8 model (442 MB) is already less than the LLC size (504 MB) on the platform. So, ideally, the weights $W$ of INT8 GEMM ($Y = X \cdot W^\top$) can reside in LLC completely. However, $X$ and $Y$ have to be read from and written to main memory, because $W$ takes up $87.7\%$ of LLC and $X$ and $Y$ data are changing at runtime. Therefore, we also measured the LLC read bandwidth to be about 1000 GB/s and we estimate the effective memory bandwidth $B_{\text{eff}}$ for INT8 GEMM on Granite Rapids according to the sizes of $X$, $W$, and $Y$ for each shape by
\begin{equation}
  B_{\text{eff}} = \frac{D_X + D_W + D_Y}{(D_X + D_Y) / B_{\text{mem}} + D_W / B_{\text{LLC}}}
\end{equation}
where $D_X$, $D_W$, and $D_Y$ are the total data sizes (bytes) of $X$, $W$, and $Y$ matrices respectively, $B_{\text{mem}}$ is the measured memory bandwidth, and $B_{\text{LLC}}$ is the measured LLC read bandwidth. The effective memory bandwidth for INT8 GEMM on Granite Rapids and the BP for each shape are listed in Table~\ref{tab:gnr-bp-eff}. After adjustment, the INT8 GEMMs are still memory-bound on Granite Rapids, but the effective memory bandwidth is improved and the balance point is reduced.

\begin{table}
  \centering
  \caption{Effective memory bandwidth and balance point for INT8 GEMM on Granite Rapids.}
  \label{tab:gnr-bp-eff}
  \footnotesize
  \begin{tabular}{cccc}
    \toprule
    GEMM Shape & Effective Memory Bandwidth (GB/s) & Arithmetic Intensity (OPs/byte) & Balance Point (OPs/byte) \\
    \midrule
    $256 \times 1024 \times 1024$ & 807.69 & 227.56 & 649.12 \\
    $256 \times 1024 \times 4096$ & 819.15 & 248.24 & 640.04 \\
    $256 \times 4096 \times 1024$ & 875.00 & 341.33 & 599.19  \\
    \bottomrule
  \end{tabular}
\end{table}

Then the estimated GEMM throughput and speedup on Granite Rapids can be computed as in Table~\ref{tab:estimated-gemm-gnr}. The aggregate throughput across mixed GEMM shapes is computed by
\begin{equation}
  \overline{P_{\text{GEMM}}} = \frac{\sum_i (counts_i \cdot F_i)}{\sum_i \left(counts_i \cdot F_i / P_i\right)}
\end{equation}
where $\overline{P_{\text{GEMM}}}$ is the aggregate GEMM throughput, $counts_i$ the occurrences, $F_i$ the number of operations, and $P_i$ the per-shape GEMM throughput, defined by
\begin{equation}
  P_i = \min(P_{\text{compute}}, \; B_i \cdot AI_i),
\end{equation}
with $B_i = B_{\text{eff},i}$ for INT8 GEMM on Granite Rapids and $B_i = B_{\text{mem}}$ otherwise.

\begin{table}[!t]
  \centering
  \caption{Estimated GEMM throughput and speedup on Granite Rapids.}
  \label{tab:estimated-gemm-gnr}
  \footnotesize
  \begin{tabular}{cccccc}
    \toprule
    GEMM Shape & \makecell{FP32 \\ Throughput \\ (TFLOPS)} & \makecell{BF16 \\ Throughput \\ (TFLOPS)} & \makecell{INT8 \\ Throughput \\ (TOPS)} & \makecell{INT8 vs FP32 \\ Speedup} & \makecell{INT8 vs BF16 \\ Speedup} \\
    \midrule
    $256 \times 1024 \times 1024$ & 19.66 & 102.40 & 183.79 & 9.35$\times$ & 1.79$\times$ \\
    $256 \times 1024 \times 4096$ & 19.66 & 114.69 & 203.35 & 10.34$\times$ & 1.77$\times$ \\
    $256 \times 4096 \times 1024$ & 19.66 & 130.33 & 298.67 & 15.19$\times$ & 2.29$\times$ \\
    Weighted Average & 19.66 & 114.69 & 218.87 & 11.13$\times$ & 1.91$\times$ \\
    \bottomrule
  \end{tabular}
\end{table}

\subsection{Estimated End-to-End Speedup Ceiling}
By applying Amdahl's law \citep{amdahl1967validity}, we estimate the end-to-end speedup ceiling for BERT-large on both platforms. For INT8 vs FP32, we use the FP32 baselines in Table~\ref{tab:fp32-baseline-hotspots}. Let $f$ be the fraction of time spent in a hotspot and $s$ be the speedup of that hotspot. The overall speedup $S$ is then:
\begin{equation}
  S = \left(\frac{1-f_{\text{GEMM}}}{s_{\text{non-GEMM}}} + \frac{f_{\text{GEMM}}}{s_{\text{GEMM}}}\right)^{-1}.
\end{equation}

The auto-mixed-precision (AMP) mechanism in PyTorch automatically converts operators to BF16 regardless of hardware support, which is a common practice in production and is used in the subsequent benchmarks. Its behaviors are as follows:
\begin{itemize}
  \item For unquantized models on Ice Lake: AMP converts inputs of all operators to BF16 but the computation remains in FP32 because Ice Lake does not support BF16. The GEMM part will be slower due to additional data type conversion, while the non-GEMM part can benefit from reduced memory bandwidth usage.
  \item For quantized models on Ice Lake: AMP converts inputs of all operators to BF16 but the computation of GEMM is in INT8 and other parts remain in FP32. The GEMM part still benefits from INT8 speedup, while the non-GEMM part can benefit from reduced memory bandwidth usage.
  \item For unquantized models on Granite Rapids: AMP converts inputs of all operators to BF16 and computation is done in BF16 (such as GEMM) or FP32 (such as elementwise operations). The GEMM and attention benefit from BF16 speedup, and the rest can benefit from reduced memory bandwidth usage.
  \item For quantized models on Granite Rapids: AMP converts inputs of all operators to BF16 and the computation of GEMM is in INT8 and other parts are in BF16 or FP32, as stated in the previous item. Thus, the non-GEMM part can benefit from BF16 computation (attention) and reduced memory bandwidth usage.
\end{itemize}

Therefore, to estimate the end-to-end speedup ceiling compared to FP32, we also need to consider the speedup of non-GEMM parts by AMP. We measured that the non-GEMM part has $1.20\times$ speedup on Ice Lake and $2.06\times$ speedup on Granite Rapids with BF16. The speedup is brought by data compression and/or native BF16 compute as mentioned above, partially offset by the overhead of data type conversion.

We use the estimated GEMM speedup and measured speedup values for the non-GEMM part. Since we only have fractions for linear-block, not pure GEMM, the estimate is approximate. Then we get the estimated end-to-end speedup ceilings of INT8 as follows:
\begin{itemize}
  \item INT8 vs FP32 on Ice Lake: $f_{\text{GEMM}} = 0.885$, $s_{\text{GEMM}} = 4.0$, $f_{\text{non-GEMM}} = 0.115$, $s_{\text{non-GEMM}} = 1.20$, $S = 3.15$.
  \item INT8 vs FP32 on Granite Rapids: $f_{\text{GEMM}} = 0.869$, $s_{\text{GEMM}} = 11.13$, $f_{\text{non-GEMM}} = 0.131$, $s_{\text{non-GEMM}} = 2.06$,  $S = 7.06$.
  \item INT8 vs BF16 on Granite Rapids: for this comparison we use the measured BF16 baseline hotspot breakdown in Table~\ref{tab:fp32-baseline-hotspots}. With $f_{\text{GEMM}} = 0.727$, $s_{\text{GEMM}} = 1.91$, $f_{\text{non-GEMM}} = 0.273$, and $s_{\text{non-GEMM}} = 1.0$, we get $S = 1.53$.
\end{itemize}

However, the realized end-to-end speedup can differ from, and is usually lower than, the estimated ceiling because
\begin{itemize}
  \item The estimated peak GEMM speedup is ideal.
  \item The quantization itself introduces additional overhead, including
    \begin{itemize}
      \item Application of smoothing factors to activations.
      \item Quantization and dequantization of tensors.
      \item Handling compensation on Ice Lake.
      \item Finding quantization parameters (scales) for dynamic quantization.
    \end{itemize}
  \item Framework overhead exists when running real workloads.
  \item The linear-block latency includes not only GEMM but also post-operations, so it is only an approximation.
  \item The LLC behavior may be more complicated than assumed.
\end{itemize}

\section{Experiments and Evaluation}

\subsection{Benchmark Setup}

The full evaluation of this work reuses the baseline collection setup from Section~\ref{sec:perf-analysis} and extends it from baseline collection to full SmoothQuant benchmarking. In particular, the same hardware platforms, sequence length, batch size, multi-instance launch strategy, AMP and AOTI deployment path are reused for throughput measurements. This section adds the model suite, quantization methods, and downstream accuracy tasks needed for the complete SmoothQuant evaluation.

We evaluate the approach in terms of both performance and accuracy using BERT-large \citep{devlin2019bert}, DistilBERT \citep{sanh2019distilbert}, and XLM-RoBERTa \citep{conneau2020unsupervised}. Minimizing accuracy loss after quantization is not the main contribution of this work, but we include it to confirm that the native SmoothQuant workflow works out of the box with little loss in quality. We use SQuAD and MultiNLI for downstream evaluation, relying on task-finetuned checkpoints. Task-finetuned checkpoints do not change performance characteristics, and according to model publishers, finetuning is required for downstream tasks.

\begin{table}[!t]
  \centering
  \caption{Models used for the performance benchmark.}
  \label{tab:models}
  \begin{tabular}{ll}
    \toprule
    Model & Model ID \\
    \midrule
    BERT-large & \texttt{bert-large-uncased} \\
    DistilBERT & \texttt{distilbert-base-uncased} \\
    XLM-RoBERTa & \texttt{xlm-roberta-base} \\
    \bottomrule
  \end{tabular}
\end{table}

\begin{table}[!t]
  \centering
  \caption{Task-finetuned checkpoints used for accuracy evaluation. The model marked by * is finetuned on SQuAD 2.0 but still workable for the test on SQuAD 1.1.}
  \label{tab:accuracy-models}
  \footnotesize
  \begin{tabularx}{\textwidth}{llY}
    \toprule
    Dataset & Model & Model ID \\
    \midrule
    SQuAD 1.1 & BERT-large & \texttt{google-bert/bert-large-uncased-whole-word-masking-finetuned-squad} \\
    SQuAD 1.1 & DistilBERT & \texttt{distilbert/distilbert-base-cased-distilled-squad} \\
    SQuAD 1.1 & XLM-RoBERTa & \texttt{deepset/xlm-roberta-base-squad2}* \\
    MultiNLI & BERT-large & \texttt{yoshitomo-matsubara/bert-large-uncased-mnli} \\
    MultiNLI & DistilBERT & \texttt{typeform/distilbert-base-uncased-mnli} \\
    MultiNLI & XLM-RoBERTa & \texttt{symanto/xlm-roberta-base-snli-mnli-anli-xnli} \\
    \bottomrule
  \end{tabularx}
\end{table}

\begin{table}[t]
  \centering
  \caption{Quantization methods used in the experiments.}
  \label{tab:quantization}
  \begin{tabular}{ll}
    \toprule
    Method & Short Name \\
    \midrule
    No quantization, no AMP & FP32 \\
    No quantization, with AMP & BF16 \\
    SmoothQuant, static quantization & Smooth-Static \\
    SmoothQuant, dynamic quantization & Smooth-Dynamic \\
    \bottomrule
  \end{tabular}
\end{table}

For SmoothQuant throughput benchmarks, we reuse the Section~\ref{sec:perf-analysis} baseline configuration and replace the baseline precision modes with the quantized configurations in Table~\ref{tab:quantization}. Both static and dynamic SmoothQuant are evaluated. For accuracy evaluation, we use the task-finetuned checkpoints in Table~\ref{tab:accuracy-models} with real downstream datasets instead of synthetic inputs. The granularity of quantization of activations and weights can be per-tensor and per-row (per-channel), and we use per-row quantization for weights in all experiments and per-tensor quantization for activations for static quantization and per-row for dynamic quantization. Performance can be improved more with per-tensor quantization of weights at the cost of accuracy loss and it is not included in the experiments. The benchmark implementation is available at \url{https://github.com/Xia-Weiwen/torchao_smoothquant_benchmark}.

\subsection{Performance Results}

The performance results on both platforms are shown in Tables~\ref{tab:perf-icx} and~\ref{tab:perf-gnr}. Overall, the proposed workflow delivers substantial end-to-end performance improvements across all evaluated models and hardware platforms. On Ice Lake, it improves throughput by approximately $1.9\times$ to $2.6\times$ over FP32 and reduces linear-block latency by more than $3\times$. On Granite Rapids, throughput speedup ranges from $4.2\times$ to $5.8\times$, and linear-block latency speedup is more than $7.5\times$, reflecting the much higher INT8 GEMM ceiling enabled by AMX. Static SmoothQuant consistently outperforms the dynamic variant in end-to-end throughput because more work is moved offline into calibration and precomputation. The comparison of INT8 vs BF16 on Granite Rapids is done in Table~\ref{tab:int8-over-bf16-gnr}. The realized speedup of INT8 over BF16 is $0.98\times$ to $1.59\times$ in throughput and $1.45\times$ to $1.84\times$ in linear-block latency. The $<1\times$ throughput speedup for DistilBERT and XLM-RoBERTa occurs because dynamic quantization overhead is higher than the GEMM speedup in these cases.

\begin{table}[t]
  \centering
  \caption{INT8 speedup over FP32 on Ice Lake.}
  \label{tab:perf-icx}
  \footnotesize
  \begin{tabular}{llrr}
    \toprule
    Model & Method & Throughput Speedup & Linear-Block Speedup \\
    \midrule
    BERT-large & Smooth-Dynamic & 2.22$\times$ & 3.59$\times$ \\
    BERT-large & Smooth-Static & 2.58$\times$ & 3.87$\times$ \\
    DistilBERT & Smooth-Dynamic & 1.90$\times$ & 3.31$\times$ \\
    DistilBERT & Smooth-Static & 2.30$\times$ & 3.35$\times$ \\
    XLM-RoBERTa & Smooth-Dynamic & 1.86$\times$ & 3.05$\times$ \\
    XLM-RoBERTa & Smooth-Static & 2.17$\times$ & 3.71$\times$ \\
    \bottomrule
  \end{tabular}
\end{table}

\begin{table}[!t]
  \centering
  \caption{INT8 speedup over FP32 on Granite Rapids.}
  \label{tab:perf-gnr}
  \footnotesize
  \begin{tabular}{llrr}
    \toprule
    Model & Method & Throughput Speedup & Linear-Block Speedup \\
    \midrule
    BERT-large & Smooth-Dynamic & 5.02$\times$ & 9.92$\times$ \\
    BERT-large & Smooth-Static & 5.82$\times$ & 8.67$\times$ \\
    DistilBERT & Smooth-Dynamic & 4.24$\times$ & 7.60$\times$ \\
    DistilBERT & Smooth-Static & 5.37$\times$ & 8.84$\times$ \\
    XLM-RoBERTa & Smooth-Dynamic & 4.17$\times$ & 7.58$\times$ \\
    XLM-RoBERTa & Smooth-Static & 5.41$\times$ & 8.71$\times$ \\
    \bottomrule
  \end{tabular}
\end{table}

\begin{table}[!t]
  \centering
  \caption{INT8 speedup over BF16 on Granite Rapids.}
  \label{tab:int8-over-bf16-gnr}
  \footnotesize
  \begin{tabular}{llrr}
    \toprule
    Model & Method & Throughput Speedup & Linear-Block Speedup \\
    \midrule
    BERT-large & Smooth-Dynamic & 1.37$\times$ & 1.84$\times$ \\
    BERT-large & Smooth-Static & 1.59$\times$ & 1.61$\times$ \\
    DistilBERT & Smooth-Dynamic & 0.98$\times$ & 1.45$\times$ \\
    DistilBERT & Smooth-Static & 1.24$\times$ & 1.69$\times$ \\
    XLM-RoBERTa & Smooth-Dynamic & 0.99$\times$ & 1.50$\times$ \\
    XLM-RoBERTa & Smooth-Static & 1.29$\times$ & 1.67$\times$ \\
    \bottomrule
  \end{tabular}
\end{table}

\subsection{Comparing Realized Speedup to Estimated Ceiling}

We compare the realized speedup of the focused BERT-large to estimated ceilings in Section~\ref{sec:perf-analysis}, summarized as Table~\ref{tab:realized-over-estimated}.

\begin{table}[!t]
  \centering
  \caption{Realized speedup of BERT-large over estimated speedup ceiling on both evaluated platforms.}
  \label{tab:realized-over-estimated}
  \footnotesize
  \begin{tabular}{lllrrrrrr}
    \toprule
     & & & \multicolumn{3}{c}{Throughput Speedup} & \multicolumn{3}{c}{Linear-Block Speedup}\\
    \cmidrule(lr){4-6}
    \cmidrule(lr){7-9}
    Platform & Method & Baseline & Realized & Estimated & Gap & Realized & Estimated & Gap \\
    \midrule
    Ice Lake & Smooth-Dynamic & FP32 & 2.22$\times$ & 3.15$\times$ & 29.59\% & 3.59$\times$ & 4.00$\times$ & 10.22\% \\
    Ice Lake & Smooth-Static & FP32 & 2.58$\times$ & 3.15$\times$ & 18.17\% & 3.87$\times$ & 4.00$\times$ & 3.21\% \\
    Granite Rapids & Smooth-Dynamic & FP32 & 5.02$\times$ & 7.06$\times$ & 28.89\% & 9.92$\times$ & 11.13$\times$ & 10.89\% \\
    Granite Rapids & Smooth-Static & FP32 & 5.82$\times$ & 7.06$\times$ & 17.56\% & 8.67$\times$ & 11.13$\times$ & 22.12\% \\
    Granite Rapids & Smooth-Dynamic & BF16 & 1.37$\times$ & 1.53$\times$ & 10.41\% & 1.84$\times$ & 1.91$\times$ & 3.58\% \\
    Granite Rapids & Smooth-Static & BF16 & 1.59$\times$ & 1.53$\times$ & -3.98\% & 1.61$\times$ & 1.91$\times$ & 15.64\% \\
    \bottomrule
  \end{tabular}
\end{table}

On Ice Lake, the throughput speedup is $2.22\times$ for dynamic quantization, about $70\%$ of the estimated ceiling, and $2.58\times$ for static quantization, about $82\%$ of the estimated ceiling. The linear-block latency speedup is $3.59\times$ for dynamic quantization, about $90\%$ of the estimated ceiling, and $3.87\times$ for static quantization, about $97\%$ of the estimated ceiling. Here, linear-block latency includes GEMM together with fused scale handling and surrounding post-operations, so it is a practical proxy instead of a strict GEMM-only metric. Because TorchInductor fusion differs across data types and models, linear-block speedup is not a strict like-for-like kernel comparison and should be interpreted as an approximation of pure GEMM speedup.

It can be seen that the realized speedup on Ice Lake is lower than the estimated ceiling, which is consistent with the analysis in Section~\ref{sec:perf-analysis}. However, the gap in linear-block latency is relatively small because Ice Lake is compute-bound for INT8 GEMM and the quantization overhead has limited impact. Thus the speedup brought by AVX512\_VNNI is easily realized.

On Granite Rapids, the throughput speedup over FP32 is $5.02\times$ for dynamic quantization, roughly $71\%$ of the estimated ceiling, and $5.82\times$ for static quantization, roughly $82\%$ of the estimated ceiling. The linear-block latency speedup is about $8.7\times$ for static quantization and $9.9\times$ for dynamic quantization, achieving roughly $78\%$ and $89\%$ of the estimated ceiling, respectively. The gap is larger for static quantization because the linear-block fusions are different for the two cases, where the heavy post-operation \verb`GeLU` is fused for static quantization but not for dynamic quantization.

When comparing INT8 to BF16, throughput speedup is $1.37\times$ for dynamic quantization and $1.59\times$ for static quantization, achieving roughly $90\%$ and $104\%$ of the estimated ceiling, respectively. The overshoot indicates the limitation of the roofline model and does not reflect a true breakthrough of the estimated ceiling. The linear-block latency speedup is $1.84\times$ for dynamic quantization and $1.61\times$ for static quantization, achieving roughly $96\%$ and $84\%$ of the estimated ceiling, respectively. Dynamic quantization has a smaller linear-block-latency gap for the same reason as above.

The gap with estimated ceilings is mainly brought by the overhead of quantization and dequantization, and searching for quantization parameters for dynamic quantization, which is more significant on Granite Rapids because INT8 GEMM is memory-bound. Again, the speedup of linear-block latency is only an approximation of GEMM speedup due to the fusion behavior.

Taking these factors into account, the overall achieved speedups are consistent with the expectation from the ceiling analysis.

\subsection{Accuracy Results}

The accuracy results in Table~\ref{tab:accuracy} show that the proposed workflow introduces negligible accuracy loss ($< 1\%$) while substantially improving performance. In the few cases where a quantized result is marginally higher than the baseline, the difference is small enough to be attributed to routine evaluation variance or mild overfitting rather than a systematic improvement from quantization itself. The results confirm that the native SmoothQuant workflow works out of the box with little loss in quality.

\begin{table}[!t]
  \centering
  \caption{Accuracy results on Granite Rapids.}
  \label{tab:accuracy}
  \scriptsize
  \resizebox{\textwidth}{!}{
  \begin{tabular}{llrrrrrrrr}
    \toprule
    Model & Metric & FP32 & BF16 & Smooth-Static & Alpha & Loss & Smooth-Dynamic & Alpha & Loss \\
    \midrule
    BERT-large & SQuAD F1 & 92.64 & 92.67 & 92.66 & 0.6 & 0.0\% & 92.63 & 0.6 & 0.0\% \\
    BERT-large & SQuAD EM & 86.41 & 86.45 & 86.49 & 0.6 & -0.1\% & 86.44 & 0.6 & 0.0\% \\
    BERT-large & MultiNLI Accuracy & 64.55 & 64.58 & 64.73 & 0.7 & -0.3\% & 64.52 & 0.6 & 0.0\% \\
    DistilBERT & SQuAD F1 & 86.39 & 86.41 & 86.00 & 0.7 & 0.5\% & 86.40 & 0.6 & 0.0\% \\
    DistilBERT & SQuAD EM & 79.02 & 79.04 & 78.49 & 0.7 & 0.7\% & 79.04 & 0.6 & 0.0\% \\
    DistilBERT & MultiNLI Accuracy & 82.21 & 82.16 & 82.26 & 0.8 & -0.1\% & 82.17 & 0.6 & 0.0\% \\
    XLM-RoBERTa & SQuAD F1 & 76.43 & 76.34 & 76.47 & 0.55 & -0.1\% & 76.59 & 0.6 & -0.2\% \\
    XLM-RoBERTa & SQuAD EM & 69.95 & 69.89 & 70.04 & 0.55 & -0.1\% & 70.30 & 0.6 & -0.5\% \\
    XLM-RoBERTa & MultiNLI Accuracy & 82.26 & 82.23 & 82.31 & 0.7 & -0.1\% & 82.26 & 0.6 & 0.0\% \\
    \bottomrule
  \end{tabular}}
\end{table}

\section{Conclusion}

We integrated and deeply optimized a SmoothQuant-based INT8 inference workflow of BERT-family NLP models in the native PyTorch stack for Intel Xeon CPUs. The work extends TorchAO's quantization workflow for SmoothQuant, expands TorchInductor's graph fusion capability to remove runtime overheads such as layout conversion and separate post-operations, and adds AVX512\_VNNI and AMX template-based GEMM kernels that participate in backend kernel selection. We further support Ice Lake through an $s8s8 \rightarrow u8s8$ transformation with precomputed compensation. A comprehensive performance analysis is conducted with roofline models for the realized speedups to demonstrate the effectiveness of the proposed workflow and optimizations. Across representative BERT-family models, the resulting system achieves near-FP32 accuracy with large throughput gains, providing an out-of-the-box path for efficient CPU deployment in native PyTorch.

\section*{Acknowledgements}

We thank Mingfei Ma, Lifeng Wang, Guobing Chen and all other colleagues at Intel Corporation who provided their support and feedback for this work.

\FloatBarrier
\bibliographystyle{unsrtnat}
\bibliography{references}

\end{document}